\documentclass[letterpaper, 10pt, conference]{ieeeconf}
\IEEEoverridecommandlockouts
\usepackage[utf8]{inputenc}

\usepackage{comment}
\usepackage{xspace}
\usepackage{graphicx}
\usepackage{overpic}
\usepackage{svg}
\usepackage{xcolor}
\usepackage{bm}
\usepackage[hang,flushmargin]{footmisc}
\usepackage{booktabs}
\usepackage{tabularx}

\usepackage{tabularx}
\usepackage{multirow}
\usepackage{booktabs}
\usepackage{colortbl}
\usepackage{float}
\usepackage{placeins}

\usepackage{array}
\newcolumntype{P}[1]{>{\centering\arraybackslash}p{#1}}  %

\usepackage{etoolbox}
\makeatletter
\patchcmd{\@makecaption}
{\scshape}
{}
{}
{}
\newcommand{\onetagright}{\tagsleft@false}
\makeatother

\usepackage{amsmath}
\usepackage{amssymb}
\usepackage{amsthm}
\usepackage{siunitx}

\newtheoremstyle{main}
{1em}                                                %
{1em}                                                %
{\itshape}                                        %
{0pt}                                                %
{\scshape}                                           %
{\\*}                                                %
{2pt}                                                %
{\thmname{#1}\thmnumber{ #2}: \thmnote{\itshape #3}} %

\usepackage{environ}

\usepackage[linesnumbered,ruled,noend]{algorithm2e}
\newcommand{\removelatexerror}{\let\@latex@error\@gobble}
\usepackage{listings}

\definecolor{gk}{RGB}{109, 109, 109}
\definecolor{gg}{HTML}{5f9411}
\definecolor{gb}{HTML}{417598}
\definecolor{gr}{HTML}{d15120}
\definecolor{gy}{HTML}{d2ad00}

\usepackage[
	activate   = {true},
	protrusion = false,
	expansion  = true,
	kerning    = true,
	spacing    = true,
	tracking   = false,
	auto       = true,
	selected   = true,
	factor     = 2000,
	stretch    = 50,
	shrink     = 20,
]{microtype}

\usepackage{csquotes}
\usepackage[
	maxbibnames=99,
	maxcitenames=2,
	natbib=true,
	style=numeric-comp,
	backend=biber,
	sorting=none,
	giveninits=true,
	url=false,
	doi=false,
	eprint=false,
	isbn=false,
]{biblatex}

\definecolor{purduegold}{HTML}{C28E0E} %

\makeatletter
\let\NAT@parse\undefined
\makeatother
\usepackage[pdfa,colorlinks,bookmarksopen,bookmarksnumbered,allcolors=purduegold]{hyperref}
\usepackage[english]{babel}

\usepackage[nameinlink,capitalise]{cleveref}
\crefname{line}{line}{lines}
\crefname{figure}{Fig.}{Figs.}
\Crefname{figure}{Fig.}{Figs.}
\crefname{equation}{Eq.}{Eqs.}
\Crefname{equation}{Eq.}{Eqs.}
\crefname{section}{Sec.}{Secs.}
\Crefname{section}{Sec.}{Secs.}
\crefname{definition}{Def.}{Defs.}
\Crefname{definition}{Def.}{Defs.}
\crefname{algorithm}{Alg.}{Algs.}
\Crefname{algorithm}{Alg.}{Algs.}
\crefname{assumption}{Asm.}{Asms.}
\Crefname{assumption}{Asm.}{Asms.}
\crefname{subassumption}{Asm.}{Asms.}
\Crefname{subassumption}{Asm.}{Asms.}
\Crefname{problem}{Problem}{Problems}
\crefname{problem}{Problem}{Problems}

\let\labelindent\relax
\usepackage[inline]{enumitem}
\usepackage{mathtools}

\graphicspath{ {./images/} }

\title{\LARGE \bf
\texttt{foam}: A Tool for Spherical Approximation of Robot Geometry
}
\author{Sai Coumar, Gilbert Chang, Nihar Kodkani, and Zachary Kingston
\thanks{
SC, GC, NK, and ZK are with the \href{https://commalab.org/}{CoMMA Lab}, Department of Computer Science, Purdue University, West Lafayette, IN, USA\newline
{\tt \{scoumar, chang940, nkodkani, zkingston\}@purdue.edu}
}%
}

\begin{document}
\maketitle

\begin{abstract}
Many applications in robotics require primitive spherical geometry, especially in cases where efficient distance queries are necessary.
Manual creation of spherical models is time-consuming and prone to errors.
This paper presents \texttt{foam}, a tool to generate spherical approximations of robot geometry from an input Universal Robot Description Format (URDF) file.
\texttt{foam} provides a robust preprocessing pipeline to handle mesh defects and a number of configuration parameters to control the level and approximation of the spherization, and generates an output URDF with collision geometry specified only by spheres.
We demonstrate \texttt{foam} on a number of standard robot models on common tasks, and demonstrate improved collision checking and distance query performance with only a minor loss in fidelity compared to the true collision geometry.
We release our tool as an open source Python library and containerized command-line application to facilitate adoption across the robotics community.
\end{abstract}

\section{Introduction}
\label{sec:introduction}

Efficient collision detection (both as binary classification and measuring distance to collision) is a fundamental challenge throughout robotics; algorithms and control strategies from real-time model-predictive control~\cite{chiu2022collision,gaertner2021collision}, sampling-based motion planning~\cite{sundaralingam2023curobo,thomason2024motions}, simulation~\cite{Coumans2016,todorov2012mujoco}, and more all rely on effective collision detection of the robot's geometry with that of the environments.
Although effective collision detection strategies exist for complex geometries (e.g., space decompositions, parallel approaches, GPU-accelerated algorithms), these approaches scale on the complexity of the involved geometry, and performance improves when the representative geometry of the robot and environment is simple.

Moreover, many robot descriptions (usually in URDF) are exported from CAD tools~\cite{tola2024understanding}, which may contain extraneous details for coarse collision checking.
These complex meshes also often contain non-manifold geometry (that is, non-watertight) and may contain geometry that violates assumptions made by geometric processing pipelines, e.g., flat planes or single faces.
A common approach is to provide simplified geometry to represent the collision geometry of the robot, either with a set of convex meshes~\cite{schulman2014motion} or with a set of primitives~\cite{zucker2013chomp}.
This is advantageous as collision checking with primitive approximations is more efficient for collision checking and signed distance queries (e.g., as in~\citet{mukadam2018continuous,sundaralingam2023curobo}), which are prevalent in modern robotics applications.
However, although these primitive approximations are useful and widely used, existing tools lack robust automation for converting existing robot geometry into these primitive decompositions.

\begin{figure}[t]
    \centering
    \begin{tabular}{ccc}
        \includegraphics[width=1\linewidth]{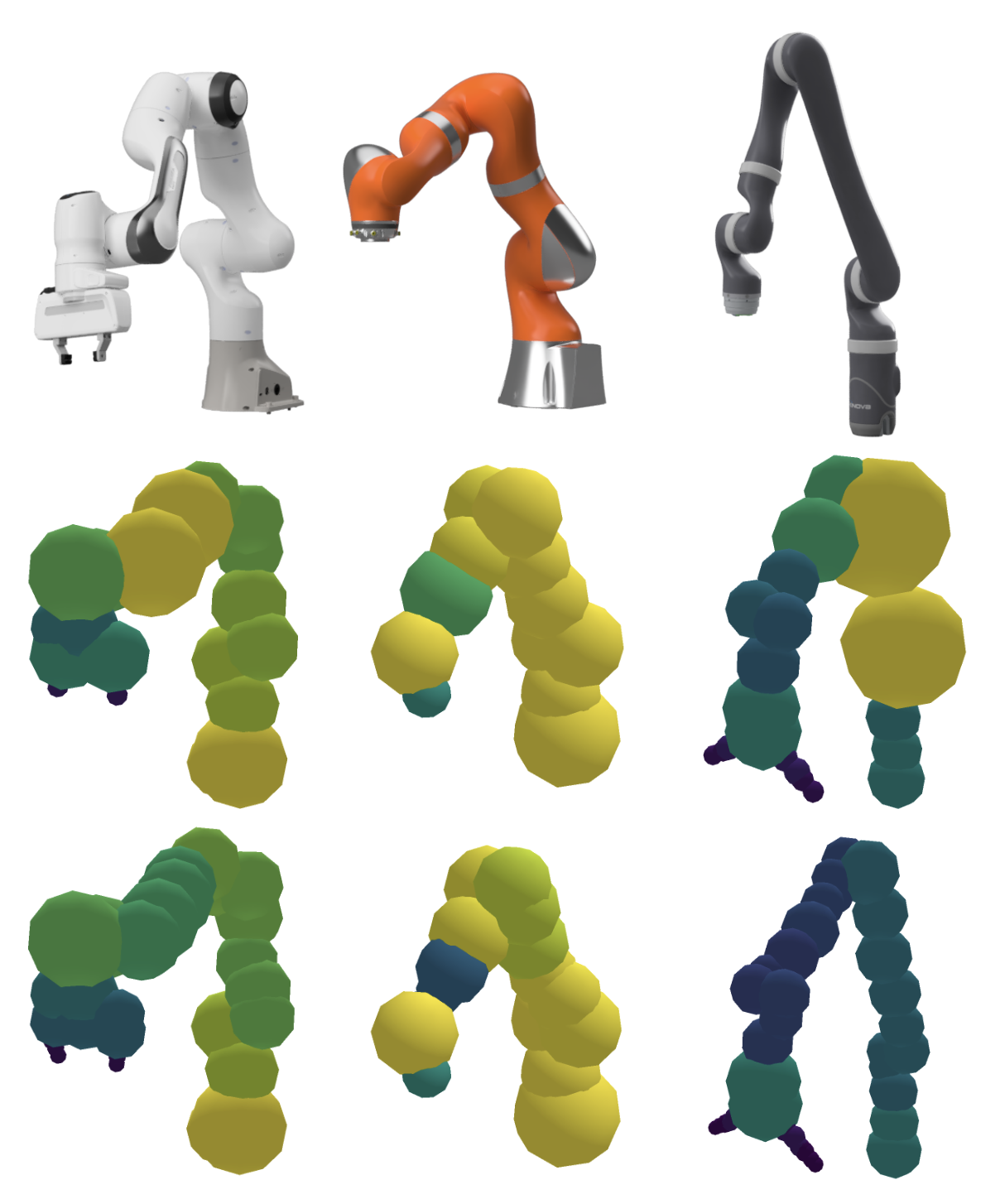}
    \end{tabular}
    \caption{Results of spherizing commonly used robots with \texttt{foam}. From left to right: Franka Emika Panda, KUKA IIWA, and Kinova Jaco robots (top row), alongside spherized approximations (middle and bottom row).
    Displayed are spherizations of the Panda with 22 and 28 spheres, the IIWA with 13 and 19 spheres, and the Kinova with 22 and 39 spheres.
    Our \texttt{foam} library automatically processes robots specified with a URDF and creates a URDF with spherical approximations of the robot's collision geometry.}
    \label{fig:spheres}
\end{figure}

This paper presents \texttt{foam}, an automated framework for generating spherical approximations of robot geometries directly from Universal Robot Description Format (URDF) specifications.
The \texttt{foam} spherization pipeline balances geometric accuracy with computational efficiency through a number of configurable approximation parameters, building upon existing work on the medial sphere-tree approximation for meshes~\cite{bradshaw2004adaptive}.
Our preprocessing pipeline handles and fixes common mesh defects gracefully, while maintaining critical geometric features needed for collision detection.
The output of \texttt{foam} is an equivalent URDF of the input robot, but with all collision geometry replaced by spheres.
We evaluated the output URDFs of \texttt{foam} across multiple standard robots and applications, and we demonstrated that \texttt{foam}-generated models achieve collision checking performance improvements while maintaining geometric fidelity suitable for practical robotic tasks.
To the authors' knowledge, while spherization tools for just meshes exist open source, nothing has been packaged as a convenient tool for roboticists;
we release our tool as open source at \url{https://github.com/CoMMALab/foam/} as a Python library and containerized command-line tool.

\section{Related Work}
\label{sec:related}
\texttt{foam} focuses on generating the spherical decomposition of a given robot geometry.
Sphere-based modeling of robots has become increasingly prominent, as spherical approximations facilitate both efficient collision checking and efficient computation of Euclidean distance transforms (EDTs)~\cite{felzenszwalb2012distance} and signed distance fields (SDFs)~\cite{ye1988signed, oleynikova2016signed, ortiz2022isdf} with respect to a robot's current configuration, which are both essential for many gradient-based and cost-map methods in robotics.
For example, in motion planning alone, trajectory optimization approaches such as CHOMP~\cite{zucker2013chomp}, STOMP~\cite{kalakrishnan2011stomp}, motion planning as inference methods such as GPMP2~\cite{dong2016motion, mukadam2018continuous}, and cuRobo~\cite{sundaralingam2023curobo} all rely on spherical models for SDF computation.
The recently proposed SIMD-accelerated sampling-based planning methods in~\citet{thomason2024motions} also rely on spherical decompositions of the robot's geometry for efficient collision detection.

Our tool is built upon existing methods for automatic approximation of a mesh by a set of spheres.
Early approaches used simple voxelizations (which are still used in, e.g.,~\citet{sundaralingam2023curobo}) or octree decompositions~\cite{hubbard1993interactive}.
Other tools fit spheres along the medial axis of the geometry~\cite{hubbard1996approximating}, that is, along the set of points inside the mesh having more than one closest point on the mesh's boundary.
For robots with simple collision geometry, e.g., cylinders, this is simple and can be done analytically, as done by~\citet{volz2018computation}.
\texttt{foam} is built on top of an improvement to these medial axis approaches, the Adaptive Medial Axis Approximation (AMAA) method of~\cite{bradshaw2004adaptive}.
Although subsequent approaches have refined the AMAA approach (e.g.,~\cite{wang2006variational}), open source code was not available to build on.
Sphere decompositions have also been used in discrete element models for particle simulations~\cite{angelidakis2021clump}.

Given the impact of mesh complexity on the performance of collision checking, rendering, and other geometric algorithms~\cite{rtcd,kockara2007collision}, there has been much research on a variety of decomposition, simplification, and approximation methods for the processing of geometry.
While alternative primitive representations have been explored in the literature, such as cuboid decompositions~\cite{sun2019learning, yang2021unsupervised} and quadric-based models~\cite{thiery2013sphere}, these have seen less adoption due to their complexity and lack of benefit for computing SDF and EDT functions.
Convex decomposition methods, e.g., V-HACD~\cite{mamou2009simple} and CoACD~\cite{wei2022approximate}, decompose a non-convex mesh into a set of convex meshes that approximate the original.
Highlighting the importance of geometry preprocessing, these decomposition techniques are critical not only for efficient geometric representation, but also for modern dataset curation.
This is evidenced by recent works that incorporate CoACD into large-scale robotics datasets~\cite{gu2023maniskill2, wang2023dexgraspnet}.
There are also other mesh simplification and low-poly approximation pipelines~\cite{cohen1996simplification, garland1997surface, calderon2017bounding, chen2023robust} that reduce the number of vertices in a mesh, but do not change the mesh into convex or primitive geometry.

\section{The Foam Library}
\label{sec:method}

The \texttt{foam} Library provides an automated framework for creating spherical approximations of robot geometries directly from Universal Robot Description Format (URDF) specifications. By converting complex polygonal meshes into simplified sphere-based representations, \texttt{foam} significantly improves computational efficiency for collision detection while maintaining sufficient geometric fidelity for practical robotics applications. The library offers a configurable pipeline that handles common mesh defects gracefully, preserves critical geometric features, and outputs equivalent URDFs with collision geometry replaced by spheres. 

\subsection{URDF Collision Geometry}
In a robot's URDF, the collision geometry is specified by some number of \texttt{<collision/>} tags in each of the robot's \texttt{<link/>} tags.
These can either be primitives (e.g., spheres, cuboids, cylinders), or by most of the time referencing external mesh files that are composed of complex polygons.
Some robot manufacturers provide convex hulls, simplified mesh geometry, or primitive approximations for collision checking, but many are simply raw CAD file output~\cite{tola2024understanding}; in some cases the details are potentially so fine that screw holes are captured in the geometry.

\texttt{foam} creates simple, accurate spherized approximations of the geometries within a robot's URDF, either specified as meshes or as primitives.
Our tool can either output spheres in a JSON database or in standard URDF representation (that is, a copy of the original robot's URDF with all the collision geometry replaced by a set of sphere primitives).
The \texttt{foam} library provides some utility functions to load URDFs, retrieve all mesh files used in a URDF, modify a URDF to include spheres, and save the output file,  respectively  \textbf{\texttt{load\_urdf(), get\_urdf\_meshes(), set\_urdf\_spheres()}} and \textbf{\texttt{save\_urdf()}}.

We note that our simplified spherized models can be easily utilized in simulators such as MuJoCo~\cite{todorov2012mujoco}, and spherization removes the dependency on external mesh assets.
Generated spheres can be used in other applications that may benefit from simplified spherical approximations.
For example, Pinocchio, which uses FCL collision checking, benefits greatly from mesh simplification due to using a collision library that is optimized for simplified geometry~\cite{carpentier2019pinocchio, fcl}.

\subsection{Mesh Preprocessing}

\begin{figure*}
    \centering
        \includegraphics[width=1\linewidth]{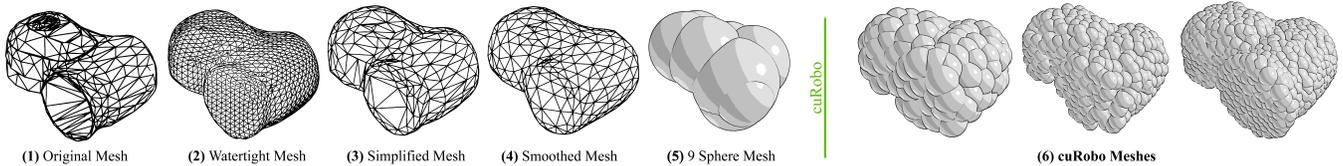}
    \caption{A representative example of a mesh being processed by \texttt{foam}.
    Here, the 6th link of the Franka Emika Panda~\cite{haddadin2024franka} is processed into a set of spheres.
    From left to right: \textbf{(1)} the original mesh is made into a \textbf{(2)} guaranteed \emph{watertight} or \emph{manifold} mesh through~\citet{huang2018robust}.
    The watertight mesh is then simplified \textbf{(3)} using a quadric-based edge collapse mesh simplification method~\cite{garland1997surface} and smoothed \textbf{(4)} with Laplacian smoothing and Humphrey filtering~\cite{vollmer1999improved}.
    Finally, spheres are fit to the processed mesh \textbf{(5)} with adaptive medial axis approximation method~\cite{bradshaw2004adaptive}.
    We also show the \textbf{(6)} spherization approach provided in cuRobo~\cite{sundaralingam2023curobo}, which voxelizes the mesh into spheres with a given sphere radius. From left to right, we show decompositions with 200 spheres and 2cm radius, 500 spheres and 1cm radius, 1000 spheres and 1cm radius}
    \label{fig:simplified}
\end{figure*}

Many polygonal mesh processing algorithms (including our desired spherization approach) assume that the input mesh is \emph{watertight} or \emph{manifold} (specifically a 2-manifold).
As implied by the name, a watertight mesh is such that it can ``hold water'', that is, it does not have holes or other broken geometry, e.g., edges with more than two incident faces, mesh faces intersecting each other, faces that are not connected together, etc.
Watertight meshes also have a clear definition of the interior and exterior of the mesh.
Unfortunately, many URDFs contain robot geometry that has been exported from automated tools~\cite{tola2024understanding}, and therefore may contain multiple defects.
These defects are also common among many meshes used in everyday tools, e.g., it is known that many of the meshes in ShapeNet~\cite{chang2015shapenet} are non-manifold.
Moreover, for our downstream tasks of collision checking and spherization, these meshes may have extraneous detail that adds additional complexity to the mesh which is unessential.

In order to handle non-watertight meshes, \texttt{foam} uses a preprocessing pipeline to handle these defects while preserving essential geometric features required for collision detection.
This preprocessing pipeline also reduces the complexity of the mesh, improving the performance of the final spherization algorithm.
A representative output from each stage of the pipeline is shown in~\cref{fig:simplified}.
Each of these steps has a number of configurable hyperparameters to control the amount of simplification, smoothing, and processing the input mesh goes through.

\begin{figure}
    \centering
    \includegraphics[width=1\linewidth]{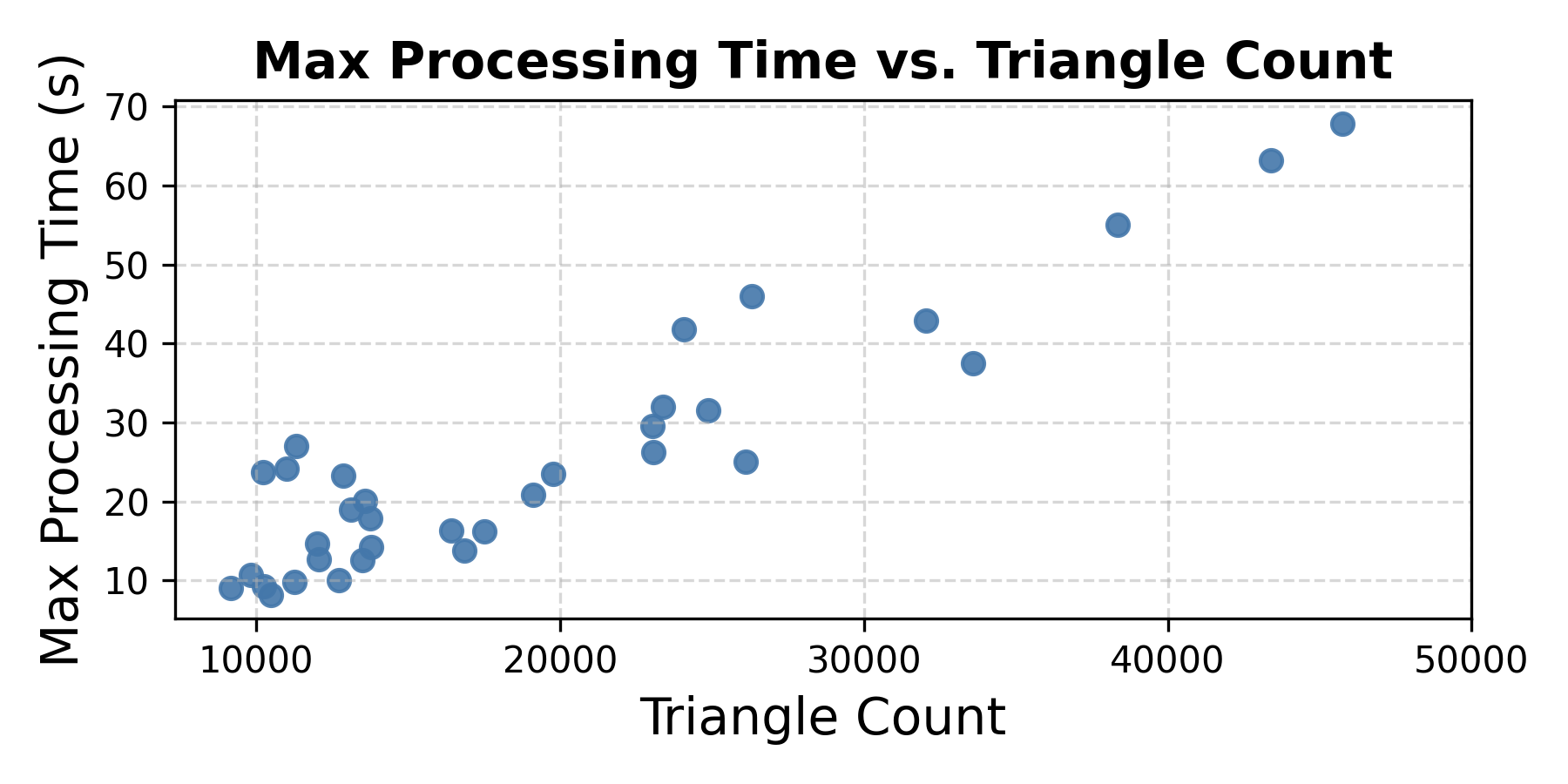}
    \caption{Time to spherize an input mesh against the number of triangles in the input mesh.}
    \label{fig:sphtime}
\end{figure}

The preprocessing workflow begins after the collision meshes are loaded with Trimesh~\cite{trimesh} (\cref{fig:simplified}-1).
The mesh is then transformed into a guaranteed watertight mesh using the method of~\citet{huang2018robust} (\cref{fig:simplified}-2).
This step is crucial and ensures topological consistency by eliminating non-manifold edges, dangling faces, and small holes that could otherwise compromise the accuracy of subsequent spherization.
Following this step, we then simplify the manifold mesh with a quadric-based edge collapse simplification method~\cite{garland1997surface} to reduce complexity, as downstream algorithms scale with the number of triangles in the mesh and are more robust with less complex input (e.g., see~\cref{fig:sphtime}).
This produces a simplified mesh (\cref{fig:simplified}-3) with significantly fewer vertices and faces, yet preserves the essential shape characteristics of the original geometry.
The process continues with a smoothing phase (\cref{fig:simplified}-4) that applies both Laplacian smoothing and Humphrey filtering~\cite{vollmer1999improved} (\textbf{\texttt{smooth\_manifold()}} in the \texttt{foam} library) to remove small-scale irregularities and noise from the mesh surface, creating a more uniform geometric representation suitable for sphere fitting.

\subsection{Spherization of Meshes}

\begin{figure*}
    \centering
    \includegraphics[width=1\linewidth]{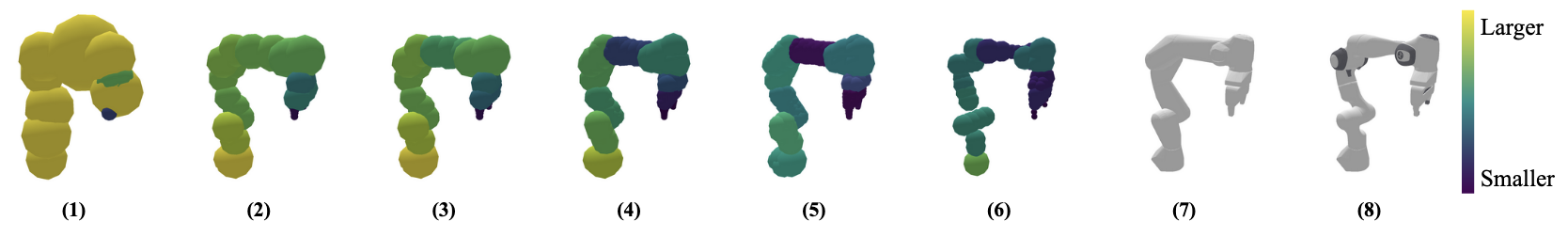}
    \caption{Franka Emika Panda Robot~\cite{haddadin2024franka} processed with \texttt{foam} to generate spherized models with increasing level of fidelity, with (left to right) \textbf{(1)} 11, \textbf{(2)} 22, \textbf{(3)} 27, \textbf{(4)} 92, and \textbf{(5)} 316 spheres.
    The spheres on the robot are colored by size, indicated by the colorbar on the far right.
    We compare the output of \texttt{foam} to \textbf{(6)} a custom-made spherization of the Panda from~\cite{fishman2023motion} with 59 spheres, which is a clear under-approximation of the robot's \textbf{(7)} true collision geometry.
    We also display \textbf{(8)}, the robot's visual geometry, as reference.}
    \label{fig:increasing}
\end{figure*}

After the mesh has been processed into a simplified, watertight mesh, it is then spherized with an implementation of the Adaptive Medial Axis Approximation (AMAA) algorithm of~\citet{bradshaw2004adaptive} (\textbf{\texttt{spherize\_mesh}} using the \textbf{\texttt{medial}} option in the \texttt{foam} library).
An output from the AMAA processing can be seen in~\cref{fig:simplified}-5, which shows how the mesh can be effectively covered with only 9 spheres.
We contrast the output of AMAA after our preprocessing to the provided spherization tool in cuRobo~\cite{sundaralingam2023curobo}, where a variety of outputs are shown in~\cref{fig:simplified}-6.
Their approach performs a simplified voxelization of the mesh into spheres with a given sphere radius.

We also expose other methods for spherization (e.g., a voxel grid-based decomposition with \textbf{\texttt{grid}}), but these notably provide less high-fidelity spherizations
There are also a number of keyword arguments to control the spherization process---a number of them are listed in the example code in~\cref{fig:code} (e.g., the maximum number of desired spheres can be specified with the \textbf{\texttt{branch}} parameter).
This control of the amount of desired spherization is illustrated in~\cref{fig:increasing}, which shows the Franka Emika Panda robot spherized with increasing amounts of spheres.
This allows users to control the fidelity and complexity of the output spherized model, either remaining coarse or reaching a close-fitting approximation.
We contrast the output of \texttt{foam} (\cref{fig:increasing}-1--5) with a custom-made spherization from~\citet{fishman2023motion} (\cref{fig:increasing}-6).
\texttt{foam}'s spherization more closely matches the true robot collision geometry (\cref{fig:increasing}-7) and is a conservative overapproximation, rather than an underapproximation.

\subsection{The Foam Tool}

\begin{figure}
\begin{lstlisting}[language=Python]
def main(mesh: str, output: str):   
    mesh_filepath = Path(mesh)
    # kwargs for spherization
    sphere_kwargs = {
        'depth': 1,
        'branch': 8,
        'method': medial,
        'testerLevels': 2,
        'numCover': 500,
        'minCover': 5,
        'initSpheres': 1000,
        'minSpheres': 200,
        'erFact': 2,
        'expand': True,
        'merge': True,
        'burst': False,
        'optimise': True,
        'maxOptLevel': 1,
        'balExcess': 0.05,
        'verify': True,
        'num_samples': 500,
        'min_samples': 1
    }
    # kwargs for mesh processing
    process_kwargs = {
        'manifold_leaves': 1000,
        'ratio': 0.2,
    }
    # Call spherize_mesh with kwargs
    spheres = spherize_mesh(
        mesh_filepath,
        spherization_kwargs=sphere_kwargs,
        process_kwargs=process_kwargs
    )
    # Write the result to a JSON file
    output = mesh_filepath.stem + "-spheres"
    with open(output + ".json", 'w') as f:
        f.write(dumps(spheres, indent=4, cls=SphereEncoder))
\end{lstlisting}
\caption{Example script to process a mesh using \texttt{foam}.}
\label{fig:code}
\end{figure}

We provide the mesh preprocessing pipeline and spherization tools as the \texttt{foam} Python library and command-line tool, along with a Docker container to provide easy use to end users.
The code is available in \url{https://github.com/CoMMALab/foam/}.
A simple example script that demonstrates how \texttt{foam} can be used as a library to convert a mesh into a set of spheres is given in~\cref{fig:code}.
After defining the \texttt{spherization\_kwargs} and \texttt{process\_kwargs}, the \textbf{\texttt{spherize\_mesh()}} function is called and the spheres are written to a JSON file.

Beyond the \texttt{foam} library, we provide standalone scripts that can be used to spherize an input mesh or URDF:\\
\textbf{\texttt{\$ python generate\_spheres.py <mesh>}}\\
\textbf{\texttt{\$ python generate\_sphere\_urdf.py <urdf>}}

\texttt{foam} has already proven useful for other publications.
For example, \texttt{foam} was used to generate the spherized geometry for the UR5, Fetch, Baxter robots as well as some environments used in~\citet{thomason2024motions,ramsey2024,quintero2024impdist}.

\section{Empirical Results}
\label{sec:results}

We observe \texttt{foam}'s effectiveness across a variety of fundamental robotics tasks in various simulators and collision checking backends. We selected MuJoCo~\cite{todorov2012mujoco}, PyBullet~\cite{Coumans2016}, and Pinocchio~\cite{carpentier2019pinocchio} as these are widely used throughout the field---MuJoCo and PyBullet are popular off-the-shelf robotics-focused physics simulators due to their wide functionality, high fidelity, and accessibility. Pinocchio is a library for robot kinematics and dynamics modeling and uses the FCL library for collision checking~\cite{fcl}. 
We present results in~\cref{tab:bigtabl} that show the difference in collision checking and signed distance query speed over these three libraries for five robots: the Hello Robot Stretch, the Franka Emika Panda, the KUKA IIWA, the Kinova Jaco, and the Rethink Robotics Baxter.
Examples of the spherized models for the Panda, KUKA, and Kinova are shown in~\cref{fig:spheres}.
While collision checking speed remains relatively consistent in PyBullet and MuJoCo between the spherized models and base models, signed distance query time is significantly faster in all engines.
FCL in particular is highly compatible with the spherized primitive representation, giving large performance gains over all robots except for the Stretch.

Additionally, we also investigate how model simplification can affect physics iteration time in a parallelized simulator, Genesis~\cite{Genesis}, shown in~\cref{tbl:genesis}. The demand for parallelized simulators has seen a recent uptick as parallelized training for large models has become more prevalent, requiring a large-scale high-speed simulation for training and validation---simplified models may be a way to increase performance in certain tasks.
Benchmarking the frames per second (FPS) in Genesis with 300 parallel simulations of the Franka Panda shows a 10\% speedup when using a simplified model with 30 spheres, compared to the original model, shown in~\cref{tbl:genesis}. Spherization also improved FPS across all models. These results demonstrate that spherized models outperform the original, offering better computational efficiency in applications that demand numerous parallelized environments.

All experiments were performed on a PC with a 13th Gen Intel(R) Core(TM) i7-13700K CPU, an NVIDIA GeForce RTX 4070Ti, and 32GB of system memory.

\begin{table*}[t]
    \centering
    \begin{tabularx}{\textwidth}{r||r r X| r r X | r r X}

     &
    \multicolumn{3}{c}{PyBullet} &
    \multicolumn{3}{c}{MuJoCo} &
    \multicolumn{3}{c}{Pinocchio (FCL)}\\
    Robot &
    $\Delta$ Col. (ms) &
    $\Delta$ Dis. (ms) &
    $\%$ Diff. &
    $\Delta$ Col. (ms) &
    $\Delta$ Dis. (ms) &
    $\%$ Diff. &
    $\Delta$ Col. (ms) &
    $\Delta$ Dis. (ms) &
    $\%$ Diff. \\
    \hline
    
    Stretch &
    -0.002 &
    -4.286 &
    -0.00001\% &
    2.329 &
    -96.189 &
    0.00251\% &
    0.649 &
    -2.479 &
    0.00386\% \\
    Panda &
    -0.002 &
    -1.274 &
    ~0.00000\% &
    0.011 &
    -14.785 &
    0.00006\% &
    -1.901 &
    -2.558 &
    0.00008\% \\
    KUKA&
    -0.000 &
    -1.586 &
    ~0.00000\% &
    0.005 &
    -1.985 &
    0.00003\% &
    -0.636 &
    -27.824 &
    0.00002\% \\
    Kinova &
    0.000 &
    -4.309 &
    ~0.00000\% &
    -0.005 &
    -0.233 &
    0.00002\% &
    -11.561 &
    -44.469 &
    0.00006\% \\
    Baxter &
    -0.001 &
    -6.879 &
    -0.00005\% &
    7.363 &
    -14.260 &
    0.00008\% &
    -1.431 &
    -1.045 &
    0.00014\% \\
    \end{tabularx}
    \caption{Comparison of collision checking and signed distance computation in three simulation and modeling frameworks: PyBullet~\cite{Coumans2016}, MuJoCo~\cite{todorov2012mujoco}, and Pinocchio~\cite{carpentier2019pinocchio}, which uses FCL~\cite{fcl} for collision checking. For each robot in each simulator, we evaluate the original robot URDF against the spherized URDF generated by \texttt{foam} over 100,000 queries in randomly generated environments, visualized in~\cref{fig:collenv}. We report the change in average query time for collision checking ($\Delta$ Col.) and the change in average distance query time ($\Delta$ Dis.) in milliseconds between the original and spherized model (negative meaning the spherized model is faster), and the percentage difference in collision check validity between the models (positive meaning more states are in collision with the spherized model). PyBullet implicitly convexifies all collision geometry and thus leads to the negative difference in collision, compared to the conservative but accurate spherized model generated by \texttt{foam}. We also note that sometimes the spherized model is less efficient for collision checking due to efficient spatial data structures used in collision testing, but distance queries are always faster. }
    \label{tab:bigtabl}
\end{table*}

\begin{table}
\centering
\footnotesize
    \begin{tabular}{r | l}
    Robot Model & Millions of FPS \\
    \hline
    Panda & 5.20 \\
    11 Spheres & 6.36 \\
    22 Spheres & 5.36 \\ 
    27 Spheres & 5.83\\
    30 Spheres & 5.73\\  
    \end{tabular}
    \caption{Simulation framerate in millions of frames per second for a Panda grasping task in the Genesis~\cite{Genesis} simulator. Using the spherized collision model provides around 10\% speedup over base model using the 30 spheres.}
    \label{tbl:genesis}
\end{table}

\begin{figure}
    \centering
    \includegraphics[width=1\linewidth]{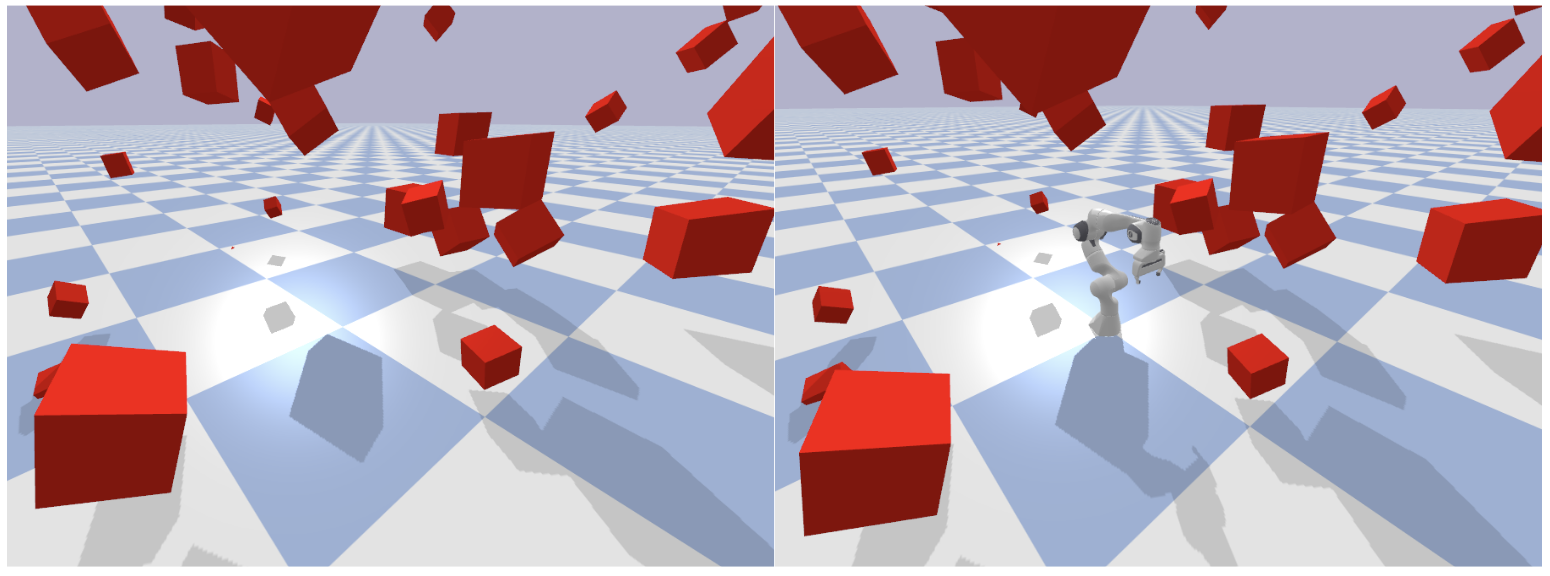}
    \caption{Left: Random environment used to evaluate collision checking, visualized in PyBullet. Right: the same environment with the Franka Emika Panda loaded inside.}
    \label{fig:collenv}
\end{figure}

\section{Discussion}
\label{sec:discussion}

We have presented \texttt{foam}, a library and tool to generate spherical approximations of robot geometry from an input URDF, addressing a critical limitation in existing tools to bridge general robots to methods that require spherical approximations.
\texttt{foam} handles common mesh defects gracefully while preserving essential geometric features; our pipeline enables roboticists to easily convert complex CAD-derived meshes into computationally efficient primitive representations.
\texttt{foam} also provides a number configurable parameters allow users to balance geometric fidelity with computational efficiency according to downstream requirements.

Our empirical results also demonstrate that spherical approximations of robot geometries provide significant computational advantages while maintaining acceptable geometric fidelity for practical robotics tasks.
Across multiple standard robots (Franka Emika Panda, Baxter, KUKA IIWA, Kinova, and Stretch), we observed substantial performance improvements in collision checking times, particularly for distance queries, with speedups of up to two orders of magnitude in some cases (e.g., Kinova through FCL-based collision checking in the Pinocchio framework).
These findings suggest that simplified robot geometry should be preferred for robotics tasks unless absolute geometric precision is essential, especially for applications involving signed distance fields or real-time control.

\texttt{foam} also opens the door for general robots to be applied to tools that require spherical models, such as cuRobo~\cite{sundaralingam2023curobo}, GPMP2~\cite{mukadam2018continuous}, VAMP~\cite{thomason2024motions}, and more.
In the future, we plan to provide not only spherical approximations in \texttt{foam}, but also convex decompositions, cuboid decompositions, and other mesh processing tools.
We are especially interested in other spherization algorithms that could be run more tightly ``in-the-loop'' to address situations that require faster spherization, such as dynamic manipulation requiring spherization of objects on the fly.

\section*{Acknowledgments}

The authors thank Andrew Lu, Carlos Quintero-Pe\~na, Pranav Jadhav, and Wil Thomason for helpful discussions and feedback.

\printbibliography{}

\end{document}